%% file: arxiv_main.tex
\documentclass[10pt,twocolumn,letterpaper]{article}

\usepackage{iccv}
\usepackage{times}
\usepackage{epsfig}
\usepackage{microtype}

\usepackage[pagebackref=true,breaklinks=true,colorlinks,citecolor=citecolor,bookmarks=false]{hyperref}

\usepackage[square,numbers,sort&compress]{natbib}

\usepackage{amsmath, amscd, amssymb, amsthm}
\usepackage{array}
\usepackage{graphicx}
\usepackage{booktabs}

\usepackage{enumitem}
\usepackage{footnote}
\usepackage{mathtools}
\usepackage{multicol}
\usepackage{multirow}

\usepackage{outlines}
\usepackage{pifont}%

\usepackage{subcaption}
\usepackage{soul}
\usepackage{tabularx}
\usepackage{tabulary}
\usepackage{url}
\usepackage{xcolor}
\usepackage{xspace}
\usepackage[utf8]{inputenc}
\usepackage{cleveref}
\usepackage{diagbox}
\usepackage{tablefootnote}
\usepackage{comment}
\usepackage{colortbl}
\usepackage{makecell}

\usepackage{tabularx}
\makeatletter
\@namedef{ver@everyshi.sty}{}
\makeatother
\usepackage{tikz,pgfplots}
\pgfplotsset{compat=1.14}
\usepackage{pgf-pie}
\usepackage{xcolor}%
\usepackage{etoolbox}
\definecolor{Azul}{rgb}{0.16, 0.32, 0.75}

\graphicspath{{./}}

\include{math_shortcuts}
\newcommand{\dataseturl}{\url{https://open-vision-language.github.io/oven}}

\iccvfinalcopy %

\def\iccvPaperID{****} %

\ificcvfinal\fi

\begin{document}

\title{Open-domain Visual Entity Recognition: \\ Towards Recognizing Millions of Wikipedia Entities}

\newcommand\blfootnote[1]{%
  \begingroup
  \renewcommand\thefootnote{}\footnote{#1}%
  \addtocounter{footnote}{-1}%
  \endgroup
}
\newcommand{\myref}[1]{\S\ref{#1}}
\newcommand{\fix}{\marginpar{FIX}}
\newcommand{\new}{\marginpar{NEW}}
\newcommand{\custompara}[1]{{\vspace{1mm}\noindent\textbf{#1}\xspace}}

\newcommand{\MultiDualEnc}{{\texttt{Mixed-DualEnc}}\xspace}
\newcommand{\MultiEncDec}{{\texttt{Mixed-EncDec}}\xspace}

\newcommand{\ourmethod}{{Our Method}\xspace}
\newcommand{\name}{\textsc{Oven}\xspace}
\newcommand{\fullname}{Open-domain Visual Entity recognitioN\xspace}

\newcommand{\datasetname}{\textsc{Oven}-Wiki\xspace}

\newcommand{\backup}[1]{{}}
\newcommand{\nlp}[1]{\texttt{\small #1}}
\newcommand{\snlp}[1]{\texttt{\small #1}}
\newcommand{\entity}[1]{\texttt{\textsc{\small #1}}}

\newcommand{\frank}[1]{{\color{brown}Frank: #1}\xspace}
\newcommand{\yi}[1]{{\color{purple}Yi: #1}\xspace}
\newcommand{\mw}[1]{{\color{blue}Ming-Wei: #1}\xspace}
\newcommand{\kristout}[1]{{\color{magenta}Kristina: #1}\xspace}
\newcommand{\uk}[1]{{\color{orange}Urvashi: #1}\xspace}
\newcommand{\kenton}[1]{{\color{cyan}Kenton: #1}\xspace}
\newcommand{\mandar}[1]{{\color{red}Mandar: #1}\xspace}

\author{
{Hexiang Hu}$^{\xspace\spadesuit}$ \quad
{Yi Luan}$^{\xspace\spadesuit}$ \quad 
{Yang Chen}$^{\xspace\spadesuit\heartsuit\dagger}$ \quad
{Urvashi Khandelwal}$^{\xspace\spadesuit}$ \\
{Mandar Joshi}$^{\xspace\spadesuit}$ \quad
{Kenton Lee}$^{\xspace\spadesuit}$ \quad
{Kristina Toutanova}$^{\xspace\spadesuit}$ \quad
{Ming-Wei Chang}$^{\xspace\spadesuit}$ \\
\small{${\spadesuit}$\xspace\textbf{Google Research} \qquad ${\heartsuit}$\xspace\textbf{Georgia Institute of Technology}}
}

\twocolumn[{%
    \renewcommand\twocolumn[1][]{#1}%
    \maketitle
    \vspace{-15mm}
    \input{tables_and_figures/intro_oven_overview}
}]

\ificcvfinal\thispagestyle{empty}\fi

\begin{abstract}
\input{sections/abstract}
\end{abstract}

\section{Introduction}
\label{sec:intro}
\blfootnote{$^\dagger$ Work was done when interned at Google Research.}
\blfootnote{{$^\ddagger$} Our dataset and evaluation toolkit is publicly available at \dataseturl}

\input{sections/intro}

\section{Open Domain Visual Entity Recognition}
\label{sec:formulation}
\input{sections/formulation}

\section{The \datasetname dataset}
\label{sec:dataset}
\input{sections/dataset}

\section{Fine-tuning Pre-trained Models for \name}
\label{sec:model}
\input{sections/model}

\section{Experiments}
\label{sec:exp}
\input{sections/experiment}

\section{Analysis}
\label{sec:analysis}
\input{sections/analysis}

\section{Related Works}
\label{sec:related}
\input{sections/related}

\section{Discussion}
\label{sec:discussion}

\input{sections/discussion}

\section*{Ethics Statement}
\label{broader_impact}

As our dataset, \ie, \datasetname, is composed of existing image recognition, image retrieval, and visual question answering datasets, we have introduced minimum risk of exposing additional social bias in our data. However, \datasetname is still at the risk of inheriting existing dataset biases. As a result, we employed existing data curation strategies~\cite{yang2020towards} to reduce such potential risks. 
Besides such risk, \datasetname also opens up new possibilities that can alleviate ethical concerns in AI systems. Specifically, \datasetname is a dataset that targets advancing research for establishing stronger grounding between the visual content and knowledge base, which can potentially contribute to building more attributed visual systems, such as a visual question answering model that produces answers based on the linked Wikipedia page, with improved interpretability and controllability. 

\section*{Acknowledgement}
We thank Boqing Gong, Soravit Changpinyo for reviewing on an early version of this paper in depth, with valuable comments and suggestions. We thank Xi Chen for providing different variants of PaLI pre-trained checkpoints. We also thank Radu Soricut, Anelia Angelova, Alan Ritter, Chao-Yuan Wu, Jiacheng Chen for discussions and feedback on the project. 

{\small
\bibliographystyle{ieee_fullname}
\bibliography{egbib}
}

\clearpage

\section{Appendix}
\input{sections/appendix}

\end{document}

%% file: math_shortcuts.tex
\DeclareMathOperator*{\argmax}{arg\,max}

\DeclareMathSymbol{@}{\mathord}{letters}{"3B}

\newcommand{\seen}{\textsc{seen}\xspace}
\newcommand{\unseen}{\textsc{unseen}\xspace}

\definecolor{navyblue}{RGB}{30,130,255}
\definecolor{citecolor}{RGB}{30,130,255}
\definecolor{lightgray}{gray}{0.9}
\definecolor{blanchedalmond}{rgb}{1.0, 0.92, 0.8}

\newcommand{\pz}{\hphantom{0}}

\usepackage{tikz}
\usetikzlibrary{shapes,snakes}

\usepackage{xcolor}
\definecolor{codegreen}{rgb}{0,0.6,0}
\definecolor{codegray}{rgb}{0.5,0.5,0.5}
\definecolor{codepurple}{rgb}{0.58,0,0.82}
\definecolor{backcolour}{rgb}{0.95,0.95,0.92}
\definecolor{darkgreen}{rgb}{0,0.4,0}
\definecolor{cerise}{rgb}{0.871, 0.192, 0.388}
\definecolor{carmine}{rgb}{0.59, 0.0, 0.09}
\definecolor{olive}{rgb}{0.332, 0.418, 0.184}
\definecolor{navyblue}{rgb}{0.496, 0.810, 0.837}
\definecolor{airforceblue}{rgb}{0.36, 0.54, 0.66}

\newcommand{\BMarker}[1]{\raisebox{0.5pt}{\tikz\fill[#1] (0,0) circle (.6ex);}}

\newcommand{\EMarker}[1]{\raisebox{0.5pt}{\tikz\draw[#1,fill=#1] (0,.6ex)--(.6ex,0)--(1.2ex,.6ex)--(.6ex,1.2ex)--(0,.6ex);}}

%% file: tables_and_figures/intro_oven_overview.tex
\begin{center}
\includegraphics[width=\textwidth]{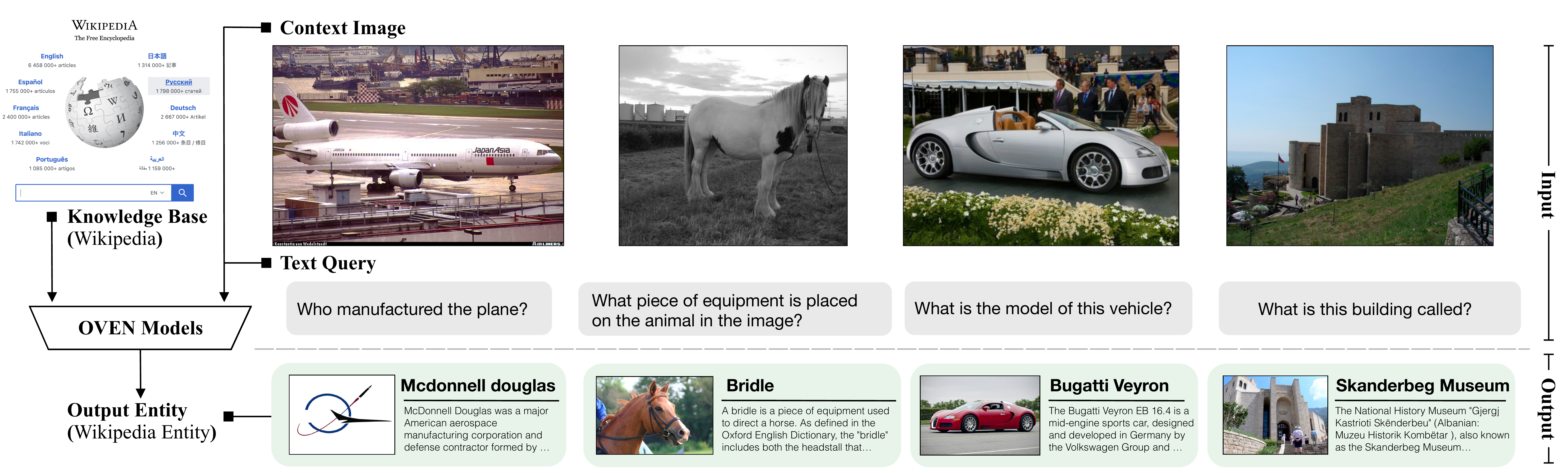}
\vspace{-7.5mm}
\captionof{figure}{
    \small
    An illustration of the proposed \name task. Examples on the right are sampled from the constructed \datasetname dataset. \name aims at recognizing entities {\em physically presented} in the image or can be {\em directly inferred} from the image.
}
\label{fig:intro_oven_overview}
\end{center}

%% file: sections/abstract.tex
Large-scale multi-modal pre-training models such as CLIP~\cite{radford2021learning} and PaLI~\cite{chen2022pali} exhibit strong generalization on various visual domains and tasks.
However, existing image classification benchmarks often evaluate recognition on a specific domain (\eg, outdoor images) or a specific task (\eg, classifying plant species), which falls short of  evaluating whether pre-trained foundational models are {\em universal} visual recognizers.
To address this, we formally present the task of \fullname (\name), where a model need to link an image onto a Wikipedia entity with respect to a text query. We construct \datasetname$^\ddagger$ by re-purposing 14 existing datasets with all labels grounded onto one single label space: Wikipedia entities. 
\datasetname challenges models to select among six million possible Wikipedia entities, making it a general visual recognition benchmark with the largest number of labels.
Our study on state-of-the-art pre-trained models reveals large headroom in generalizing to the massive-scale label space. We show that a PaLI-based auto-regressive visual recognition model performs surprisingly well, even on Wikipedia entities that have never been seen during fine-tuning. 
We also find existing pretrained models yield different strengths: while PaLI-based models obtain higher overall performance, CLIP-based models are better at recognizing tail entities. 

%% file: sections/intro.tex
Pre-trained large language models~\cite{brown2020language, chowdhery2022palm}, \textit{inter alia}, have shown strong transferable text processing and generation skills in tackling a wide variety of natural language tasks~\cite{wang2018glue,sarlin2020superglue,srivastava2022beyond} across languages and task formats, while requiring very few manually labeled per-task examples. At the same time, while there has been equally impressive progress in multi-modal pre-training~\cite{radford2021learning, chen2022pali}, it remains unclear whether similarly universal visual skills, \ie, recognizing millions of coarse-grained and fine-grained visual concepts, have emerged. {\em Are pre-trained multi-modal models capable of recognizing open-domain visual concepts?}

Answering this question requires a visual recognition dataset with broad coverage of visual domains and tasks, under a universally defined semantic space. Existing recognition benchmarks such as ImageNet ~\cite{russakovsky2015imagenet,ridnik2021imagenet}, Stanford Cars~\cite{krause2013cars196}, or SUN database~\cite{xiao2010sun} represent a large number of visual concepts, but make specific assumptions about the granularity of the target concepts (e.g. building type such as ``castle'' in ImageNet but not a specific building in the world such as ``Windsor Castle''), or limit attention to concepts of the same type such as car models/years. Visual question answering (VQA) datasets test models' abilities to recognize concepts which can be of more flexible granularities and object types, but in practice existing VQA datasets tend to focus on higher-level categories. 
We aim to assess models' abilities to recognize visual concepts from a close to universal, unified space of labels that covers nearly all visual concepts known to humankind, and at a flexible level of granularity, specified by a user or a downstream application. Given a short specification of each element in the target space of visual concepts (such as a textual description), multimodal pre-trained models could in principle recognize concepts without seeing labeled instances covering each of them. 

Towards evaluating models on such universal visual recognition abilities, we introduce the task of {\bf O}pen-domain {\bf V}isual {\bf E}ntity recognitio{\bf N} (\name), targeting a wide range of entities and entity granularities, including animals, plants, buildings, locations and much more. Particularly, we construct \datasetname by building on existing image recognition and visual QA datasets and unifying their label spaces/granularities and task formulations.  For our unified label space, we use English Wikipedia which covers millions of visual entities of various levels of granularity and also includes a specification of each entity via its Wikipedia page (containing entity name, text description, images, etc.). Wikipedia also evolves as new entities appear or become known in the world, and can be used as a first approximation of a universal visual concept space.

We re-purpose 14 existing image classification, image retrieval, and visual QA datasets, and ground all labels to Wikipedia. In addition to unifying labels, we unify input recognition intent specifications, which is necessary when combining specialized datasets with the goal of evaluating universal recognition. 
Given an image showing a car and a tree behind it, \name makes the recognition intent explicit via a natural language query such as ``What is the model of the car?'' or ``What is the species of the tree?''. Therefore, the \name task takes as input an image and a text query\footnote{A query can be expressed in different formats; in this paper, we choose to use a question to reflect the intent.} that expresses visual recognition intent with respect to the image. The goal is to provide an answer by linking to the correct entity (e.g. \entity{Bugatti Veyron} or \entity{Bactris gasipaes}) out of the millions of possible Wikipedia entities, each coming with descriptions and a relevant set of images from its Wikipedia page (see Figure~\ref{fig:intro_oven_overview}). Importantly, 
\name requires recognition of entities that were \unseen in the training data. Models can still take advantage of the text description and/or images on the Wikipedia page of the \unseen entities, as well as knowledge acquired through pre-training.

Human annotators were hired to help create \datasetname for two reasons. First, grounding labels from the component datasets into Wikipedia entities is non-trivial due to language ambiguity. For example, `\nlp{Tornado}' can be a weather phenomenon or a type of airplane (\entity{Panavia Tornado}). 
To reduce such ambiguity in the grounding, we take multiple steps to refine the labels, including the use of human annotators, a state-of-the-art textual entity linking system~\cite{de2020autoregressive}, and heavy filtering. Second, creating unambiguous textual query intents is also challenging. In many cases,  a text query 
can lead to multiple plausible answers (e.g. of various granularities), and a human often needs to make revisions to make sure no other objects could be correct answers. For our training and development/test sets we rely on semi-automatic processing, but additionally introduce a gold evaluation set, for which annotators thoroughly corrected entity linking errors and rewrote ambiguous input query intents. 

Based on \datasetname, we examine two representative multi-modal pre-trained models, PaLI~\cite{chen2022pali} and CLIP~\cite{radford2021learning}, to establish an empirical understanding of the state-of-the-art in universal entity recognition. Particularly, these two models are used for creating an auto-regressive visual entity recognition model (similar to~\cite{de2020autoregressive}) and a visual entity retrieval model, respectively.
Our study suggests that there is a large room for improvement in generalizing to the massive label space. We show that the PaLI-based auto-regressive visual recognition model performs surprisingly well, even on Wikipedia entities that have never been seen during fine-tuning. Digging deeper, we discover that CLIP variants and PaLI-based models make very different kinds of errors. Particularly, PaLI dominates in recognizing popular Wikipedia entities, whereas CLIP models can win consistently on recognizing tail entities.

%% file: sections/formulation.tex
To drive progress in universal entity recognition, we propose the task of \fullname (\name). There are two desiderata that we would like to meet for the \name task. First, there should exist a universal label space. In \name, we make use of a multi-modal knowledge base, such as Wikipedia, to serve as the universal label space, covering millions of entities. Second, the answer label for each \name input should be unambiguous. This is particularly challenging when the label space is very large and multi-granular. To accomplish this, \name makes use of input text queries to define the recognition intent (\eg, identifying car types or car models),  allowing visual concepts from different granularities to be unambiguously specified.

\input{tables_and_figures/fig_model.tex}

\custompara{Task Definition} 
The input to an \name model is an image-text pair $x = (x^p, x^t)$, with the text query $x^t$ expressing intent with respect to the corresponding image $x^p$.
Given a unified label space $\mathcal{E}$ which defines the set of all possible entities, the knowledge base $\mathcal{K}= \{(e, p(e), t(e)) \mid e \in \mathcal{E}\}$ is a set of triples, each containing an entity $e$, its corresponding text description $t(e)$ (\ie, name of the entity, description, etc.) and a (possibly empty) set of relevant images $p(e)$. 
For instance, an entity $e=\text{\entity{Q7395937}}$ would have a corresponding textual description $t(e)=\text{`\snlp{Name: Sabatia campestris; Description:$\ldots$}'}$\footnote{In this paper, we only consider using the name of the entity as its textual representation, despite the fact that more textual descriptions are available.} and a set $p(e)$ containing one or more images from the corresponding Wikipedia page\footnote{\url{https://en.wikipedia.org/wiki/File:Sabatia_campestris_Arkansas.jpg}} of \entity{Sabatia campestris}.
We consider the combination of $t(e)$ and $p(e)$ the {\em multimodal knowledge} for the entity $e$.
As \name is a recognition task, we focus on recognizing and linking entities that are {\em physically} present in the image.\footnote{Extending this framework to entities that are not physically present in the image (e.g. the inventor of the airplane) is also valid and useful. See a follow-up works~\cite{yang2023infoseek} for more details.}

The goal of learning for \name is to optimize a function $f_{\Theta}$ that predicts the entity $e$ from a given test example $x = (x^p, x^t)$ and the associated knowledge base of triples $\mathcal{K}$. There are different ways to utilize the information available in $\mathcal{K}$, and models may choose to use only a subset of this information. Figure~\ref{fig:model_comparison} presents two typical ways of modeling \name. For encoder-decoder models~\cite{chen2022pali,wang2021simvlm}, the most straight-forward utilization is to memorize the entities of the database $\mathcal{K}$ into model parameters $\Theta$ via pre-training and fine-tuning, and then {\em generate} entity names directly during inference.  Given that the generated name might not appear in the database, BM25 is used to map the prediction to the entity with the closet name in the available database
For dual-encoder models~\cite{faghri2017vse++,chen2021learning,radford2021learning,jia2021scaling}, an alternative is to explicitly compare a given test example $x$ to representations of entities $e \in \mathcal{E}$, making the prediction an \textit{entity retrieval} problem. We refer to Section~\ref{sec:model} for concrete examples of how to implement \name models.

\custompara{Data Split and Evaluation}
Due to \name's goal of evaluating pre-trained multi-modal models, we only provide a partial set of visual concepts (\ie, \seen categories) for model training or fine-tuning. For evaluation, an
\name model is tested on generalization to entities not present in the fine-tuning data (thus \unseen), without forgetting the \seen concepts. The models need to either acquire information from the knowledge base, or make a prediction using knowledge obtained during pretraining. We evaluate \name with a metric aiming to balance performance between \seen and \unseen entities using a harmonic mean, as shown below: 
\begin{equation}
    \small
    \textsc{hm}(\texttt{Acc}_{\seen}, \texttt{Acc}_{\unseen}) = 2 \; / \; (\frac{1}{\texttt{Acc}_{\seen}} + \frac{1}{\texttt{Acc}_{\unseen}}) \label{eqn:hm}
\end{equation}
Harmonic mean equally weighs the importance of the \seen and \unseen subsets, and penalizes models with a short barrel. Further details are provided in~\myref{sec:dataset}.

\custompara{\name versus recognition benchmarks}
Given that an \name model need to
generalize to \unseen entities, it is required to predict over all KB entities, which can exceed 6 million in our experiments (\eg, the size of English Wikipedia). This is orders of magnitude larger than existing benchmarks.
Second, the large label space has made the generalization to \unseen entities the most critical criterion for a successful \name model, which also allows future open-domain evaluation\footnote{One can collect and label a new set of entities from Wikipedia, to serve as a new evaluation data for \name models}. Third, \name requires models to do multi-modal reasoning, \ie, comprehending the text query within its visual context, to predict the answer entity. 

\custompara{\name versus Visual QA tasks} 
\name can be considered as a VQA task because its input format is the same as that of standard VQA models (\eg, text query + image). However, \name is specialized and focuses solely on recognition, with the text input serving mainly for intent disambiguation. Moreover, \name models are required to generate the  name of an entity that exists in a given KB (like models for text entity linking tasks), while VQA models output free-form answers (such as \nlp{yes/no} for verification questions and numbers for counting questions). 

\custompara{From \name to Knowledge-Intensive VQA}
Although this paper aims to evaluate pre-trained multi-modal models on universal visual entity recognition, we highlight that models that excel at \name can serve as foundational components for systems that can answer knowledge-intensive questions. For example, given an image and a question ``When was the church built?'', one could apply an \name model to link the image to a concrete church's Wikipedia page and then extract the answer from that document. A follow-up work has conducted a thorough study on the value of Wikipedia grounding for answering knowledge-intensive visual questions~\cite{yang2023infoseek}.

%% file: tables_and_figures/fig_model.tex
\begin{figure}[th]
    \centering
    \tabcolsep 0pt
    \begin{tabular}{cc}
        \includegraphics[width=0.245\textwidth]{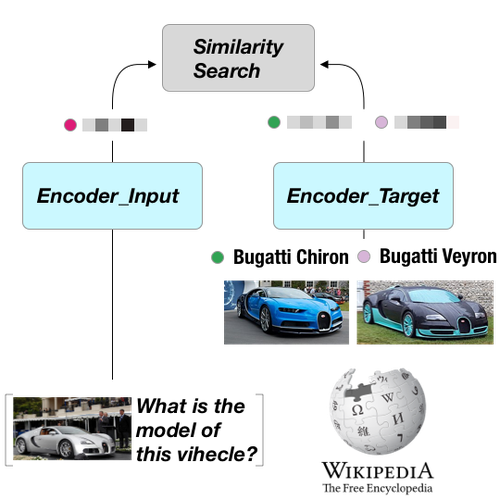} & \includegraphics[width=0.245\textwidth]{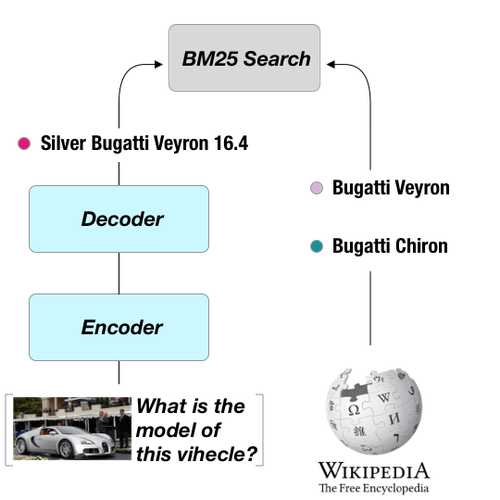} \\
        (a) Dual Encoder & (b) Encoder Decoder
    \end{tabular}
    \caption{\textbf{Illustration on two \name Models. }}
    \label{fig:model_comparison}
    \vspace{-2mm}
\end{figure}

%% file: sections/dataset.tex
\input{tables_and_figures/dataset_stats}

Based on the task formulation of \name, we create the \datasetname dataset by combining 14 existing datasets, grounding their labels to Wikipedia, resolving label ambiguities, and providing unambiguous textual query intents for all examples.
The 14 datasets were originally created for image recognition/retrieval, and visual question answering. 
Below is the complete list:
\begin{itemize}
    \setlength{\leftmargin}{0pt}
    \setlength{\itemsep}{0pt}
    \setlength{\parskip}{0pt}
    \item \textbf{Image Recognition Datasets}: ImageNet21k-{P}~\cite{russakovsky2015imagenet,ridnik2021imagenet}, iNaturalist2017~\cite{van2018inaturalist}, Cars196~\cite{krause2013cars196}, SUN397~\cite{xiao2010sun}, Food101~\cite{bossard14food}, Sports100~\cite{sports100}, Aircraft~\cite{maji13fine-grained}, Oxford Flower~\cite{nilsback2008automated}, Google Landmarks v2~\cite{weyand2020google}.
    \item \textbf{Visual QA Datasets}: VQA v2~\cite{goyal2017making}, Visual7W~\cite{zhu2016visual7w}, Visual Genome~\cite{krishna2017visual}, OK-VQA~\cite{marino2019ok}, Text-VQA~\cite{singh2019towards}. 
\end{itemize}
These datasets \backup{generally} belong to two groups: image recognition (or retrieval) which provides {\em diverse visual entities}, defined as the \textbf{Entity Split} (ES); and VQA which provides {\em visually-situated natural language queries},  defined as the \textbf{Query split} (QS). 
For examples that originate from VQA datasets, we employ human annotators to write templated rules and filter out questions that do not lead to visual entity answers that are present in the image. For examples from recognition datasets, we first extract the super-category  of their label (using the Wikipedia database), and then apply a templated query generation engine to generate a query with unambiguous intent that leads to the label (details in the Appendix).

\custompara{Label Disambiguation and Human Annotation} 
Grounding the labels of 14 datasets to Wikipedia entities is challenging, and we perform the following steps to accomplish this. We first apply a state-of-the-art textual entity linking system~\cite{de2020autoregressive} to recognize text labels and map them into Wikipedia. Human annotators are used to write rules to detect bad linking results or unlinkable labels (e.g. numbers), and correct entity linking errors. The union of original dataset labels were linked to 20,549 unique Wikipedia entities, each with a number of examples for the purpose of training and evaluation. Meanwhile, we construct the candidate label space using the English Wikipedia snapshot from \textit{Oct. 1 2022}, by removing all disambiguation, redirect, and media file pages.  As shown in Figure~\ref{fig:intro_oven_overview} (right), this left us with 6,063,945 Wikipedia entities in total. Note that we only consider using the first Infobox images~\cite{wiki_infobox_picture} from each page to serve as the visual support for each Wikipedia entity; these are available for 2,032,340 entities. 

We further perform human annotation to create a high-quality evaluation dataset. Specifically, we hired over 30 dedicated annotators to validate the entity links in \texttt{<}image, query, answer\texttt{>} triplets sampled from the test split. They were asked to re-annotate the triplets with access to the visual context, ensuring that the query leads to the correct Wikipedia entity answer. Through this process, we collected 24,867 natural language queries, equally distributed over triplets originally sampled from the Entity and Query splits (\ie, test splits). We asked the annotators to rewrite the queries so that no other object in the image could be a valid answer.  As a result,  the percentage of unique queries in the total examples (17,669 out of 24,867) as shown in Table \ref{fig:dataset_statistics} (mid) is significantly higher in the human set than in the other sets. This brings higher query generalization challenges for the human eval set. We report  results using the same evaluation metrics on the human data, with respect to \seen and \unseen entities. Figure~\ref{fig:intro_oven_overview} provides a glance at the human annotated data. 

\custompara{Dataset Statistics} 
Figure~\ref{fig:dataset_statistics} (left) presents the general distribution of the super-categories for our final collection of  Wikipedia entities that have positive examples. 
Figure~\ref{fig:dataset_statistics} (right) shows detailed statistics for queries and entities for each of the fine-tuning (train), validation, test, and human splits. Note that the models do not know which entities are present in the val/test/human set, and must scan through the whole KB to make predictions.  The \# of \seen/\unseen examples indicates the \# of examples of which the positive entity labels are in the \seen/\unseen split.  

\custompara{Evaluation Details} 
As aforementioned, we evaluate models by asking them to predict one out of over 6 million English Wikipedia entries.
While our data does not cover all 6 million labels as positive examples, models
still need to consider all possible outputs due to the presence of \unseen entities.
We measure the models' performance using both the Entity Split (ES) and Query Split (QS). Specifically, we first compute the harmonic mean of accuracy over examples from the \seen and \unseen classes, as $\texttt{Acc}_{\texttt{ES}} = \textsc{hm}(\texttt{Acc}_{\texttt{ES}\xspace\seen}, \texttt{Acc}_{\texttt{ES}\xspace\unseen})$ and $\texttt{Acc}_{\texttt{QS}} = \textsc{hm}(\texttt{Acc}_{\texttt{QS}\xspace\seen}, \texttt{Acc}_{\texttt{QS}\xspace\unseen})$ as the Equation~\ref{eqn:hm}. 
Then we further calculate the harmonic mean between splits $\textsc{hm}(\texttt{Acc}_{\texttt{ES}}, \texttt{Acc}_{\texttt{QS}})$ to reward models that do well on both splits.
We use the validation data, which contains examples from subsets of both \seen and \unseen entities, for model selection, and we measure performance on the test split and the human evaluation set.

%% file: tables_and_figures/dataset_stats.tex
\newtoggle{showpct}
\makeatletter
\patchcmd{\pgfpie@slice}%
{\scalefont{#3}\beforenumber#3\afternumber}%
{\iftoggle{showpct}{\scalefont{#3}\beforenumber#3\afternumber}{}}%
{}{}
\makeatother

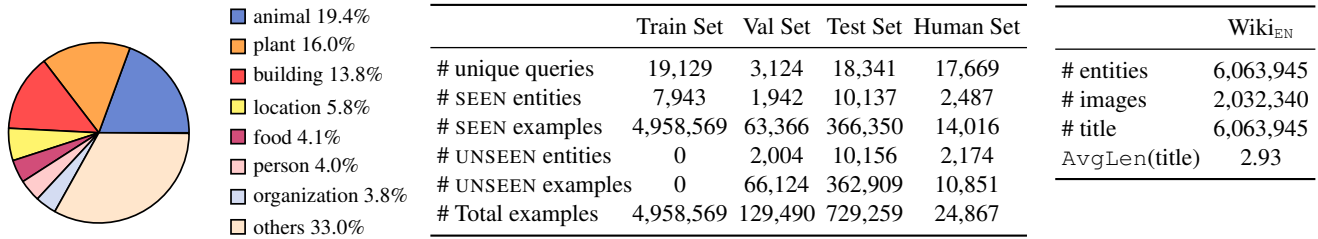
\begin{figure*}[htp]
    \begin{subfigure}[]{.3\textwidth}
        \scalebox{0.8}{
        \begin{tikzpicture}[scale=0.5]
        \pie[radius=3,text=legend,
            color={Azul!70,orange!70,red!70,yellow!70, purple!70, red!20, Azul!20, orange!20}]
            {
            19.4/animal 19.4\%,
            16.0/plant 16.0\%, 
            13.8/building 13.8\%,
            5.8/location 5.8\%,
            4.1/food 4.1\%,
            4.0/person 4.0\%,
            3.8/organization 3.8\%,
            33.0/others 33.0\%
            }
        \end{tikzpicture}
        }

    \end{subfigure}
    \hspace{8pt}
    \begin{subfigure}[]{.445\textwidth}
        \centering
        \small
        \tabcolsep 2pt
        {
        \begin{tabular}{@{\;}l@{}cccc@{\;}}
            \toprule
             &  Train Set  & Val Set & Test Set & Human Set \\ \midrule
            \# unique queries & 19,129 & 3,124 & 18,341 & 17,669 \\
            \# \seen entities & 7,943 & 1,942 & 10,137 & 2,487 \\
            \# \seen examples & 4,958,569 & 63,366 & 366,350 & 14,016 \\
            \# \unseen entities & 0 & 2,004  & 10,156 & 2,174 \\
            \# \unseen examples & 0 & 66,124 & 362,909 & 10,851\\
            \# Total examples & 4,958,569 & 129,490 & 729,259 & 24,867\\
            \bottomrule
        \end{tabular}
        }
    \end{subfigure}
    \hspace{8pt}
    \begin{subfigure}[]{.155\textwidth}
        \small
        \tabcolsep 2pt
        \raisebox{1.85cm}
        {
        \begin{tabular}{@{\;}l@{\;\;}c@{\;}}
            \toprule
             &  Wiki$_{\texttt{EN}}$ \\ \midrule
            \# entities & 6,063,945 \\
            \# images & 2,032,340 \\
            \# title & 6,063,945 \\
            \texttt{AvgLen}(title) & 2.93 \\

            \bottomrule
        \end{tabular}
        }
    \end{subfigure}
	\caption{
    	\small
	    Dataset Statistics of the \datasetname. 
	    {\bf Left: } Distribution of super-categories of entities that have positive examples (See Appendix for more details). {\bf Mid: }
	    Statistics of different splits of the \datasetname. {\bf Right: } Properties of the Wikipedia dump-\texttt{2022/10/01}.
    }
    \vspace{-5mm}
    \label{fig:dataset_statistics}
\end{figure*}

%% file: sections/model.tex
We evaluate two prominent pre-trained multi-modal models: CLIP~\cite{radford2021learning}, a widely-used dual encoder model for image and text, and PaLI~\cite{chen2022pali}, a state-of-the-art pre-trained encoder-decoder model. Figure~\ref{fig:model_comparison} has illustrated high-levelly on how encoder-decoder and dual encoder models can model the task of \name. In the following, we demonstrate with more details about how these two models can be fine-tuned for \name.

\subsection{Dual encoders: CLIP and its variants for \name}
One can naturally apply CLIP on \name by treating it as an image-to-text retrieval task. For an input image $x^p$, the image encoder is used to form an image embedding. Then the predicted entity could be retrieved by finding the entity that has the maximum dot product value between the entity text embeddings and entity image embeddings among the entire entity database. However, this naive implementation ignores the input intent $x^t$ and the entity images $p(e)$. 

In the following, we present two variants of CLIPs: CLIP Fusion and CLIP2CLIP. The goal of these two variants is to use all of the information provided in the \name task. Both variants learn a function $f_\Theta$ that maximizes the score of the target entity for the given input image-query pair, using multimodal knowledge from the knowledge base. Given a test example $x = (x^p, x^t)$ and the knowledge base of triples $\mathcal{K}$, the function is used to make a prediction,
\begin{equation}
    e' = \argmax_{e \in \mathcal{E}} f_{\Theta}(x^p, x^t, p(e), t(e))
\end{equation}

\custompara{CLIP Fusion} adopts the pre-trained CLIP model as the featurizer to develop this system, via adding a 2-layer Multi-Modal Transformer on top of the CLIP image and text features as a mixed-modality encoder. The left encoder (for an input image-query pair) and the right encoder (for multi-modal knowledge information) use the same architecture, but do not share parameters.
We fine-tune all of their parameters on the \datasetname, which includes both the pre-trained CLIP weights and randomly initialized Transformer weights.

\custompara{CLIP2CLIP} relies more heavily on the pre-trained CLIP model and introduces only a minimal set of new parameters (\ie, four) to re-weigh and combine CLIP similarity scores.
Particularly, it computes the cosine similarity between $\texttt{<}x^p, t(e)\texttt{>}$, $\texttt{<}x^t, p(e)\texttt{>}$, $\texttt{<}x^p, p(e)\texttt{>}$, and $\texttt{<}x^t, t(e)\texttt{>}$, using the image and text encoders of CLIP, respectively. Then it aggregates these similarities by multiplying them with a learnable vector that reflects importance weights.

\custompara{Scaling to 6 million candidates.}
It is expensive to perform dot product scoring with respect to 6 million webpages on-the-fly. Fortunately, there exist approximate algorithms for maximum inner product search whose running time and storage space scale sub-linearly with the number of documents \cite{mips_cone,mips_alsh,mips_binary}. In all our experiments, we use ScaNN~\cite{avq_2020} as our library for entity retrieval.

\subsection{Encoder-Decoder: PaLI for \name}
PaLI~\cite{chen2022pali} is a sequence-to-sequence model pre-trained on web text, image-text pairs (\ie, WebLI) and other sources. PaLI can accept both an image and text as input and generates text as output. In order to map the PaLI predictions to the knowledge base, we run a BM25~\cite{robertson2009probabilistic} model to retrieve the most similar Wikipedia entity name for every generated text output. %
We found that this can slightly but consistently improve the entity recognition results. Note that we directly fine-tune PaLI on the \name training data, which does not cover all entities and questions appearing in our Dev and Test splits. However, we found that PaLI is still able to handle entities that are unseen during fine-tuning due to the knowledge acquired during pre-training. To make the comparison with CLIP more comprehensive, we report results on both PaLI-3B and PaLI-17B. The former PaLI variant is at the same magnitude (in its number of parameters) as the largest CLIP model, and the latter PaLI variant is one magnitude larger, and much stronger based on other evaluation~\cite{chen2022pali}.

%% file: sections/experiment.tex
We first describe the essential experimental setups in \myref{sec:exp_setups}, and then present the main benchmark results in \myref{sec:main_results}.

\subsection{Experimental Setups}
\label{sec:exp_setups}

\custompara{Pre-trained Model Details.} For all the CLIP variants, we employ the largest CLIP checkpoints, \ie, \texttt{ViT-L14}, which leverages Vision Transformer~\cite{dosovitskiy2020image,vaswani2017attention} as its visual backbone. For the PaLI model~\cite{chen2022pali}, we make use of the 3\texttt{B} and 17\texttt{B} parameter pre-trained models provided by the original authors, for fine-tuning on \name. 

\custompara{Data Processing Details.} We process all images in our dataset by resizing them to 224$\times$224, linearize them into a sequence of 14$\times$14 patches, and apply the normalization technique consistent with each model's pretraining to preprocess the images. For natural language text, we perform tokenization based on the adopted pre-trained model's original vocabulary. For CLIP variants that encode Wikipedia images for entity retrieval, we apply the same image processing pipeline whenever the image is available. When the Wikipedia entity does not have an infobox image, we use a black image to represent the visual support. 

\subsection{Benchmark Results}
\label{sec:main_results}

\input{tables_and_figures/exp_main_val_table}
\input{tables_and_figures/exp_main_test_table}

\custompara{Main Results} Results on the validation set are presented in Table~\ref{tab:main_results_validation}, and include performance on the Entity and Query splits, as well as the overall combined scores.

There are several interesting (perhaps surprising) observations from Table~\ref{tab:main_results_validation}. First, while CLIP variants such as CLIP Fusion and CLIP2CLIP are utilizing more information from Wikipedia (\ie, entity names and entity images), they are weaker than the auto-regressive PaLI-3B and PaLI-17B model, across most evaluation data splits. This suggests that high-capacity generative multi-modal pre-trained models are capable of recognizing visual entities. Second, this performance gap is more apparent on the query split than the entity split, potentially due to the VQ2A pre-training objectives~\cite{changpinyo2022all}
and the underlying powerful language models~\cite{raffel2020exploring} employed by the PaLI model. 

Comparing all CLIP-based models, we observe that CLIP Fusion and CLIP2CLIP, which uses all Wikipedia information are generally performing better than the vanilla CLIP model, showcasing the benefits of multimodal information from Wikipedia. Meanwhile, we also observe that CLIP Fusion, where two new layers have been added on top of pretrained CLIP, shows very strong results on \seen entities for both the Entity and the Query splits, but weak results on \unseen entities, thus leading to lower overall performance. The CLIP2CLIP model, on the other hand, is capable of retaining the cross-entity generalization performance while improving its prediction accuracy on \seen entities. %

Comparing the PaLI models, we observe a drastic improvement as the number of parameters in the models increased. Particularly, PaLI-17B has a double-digit performance gain in the overall performances, against the PaLI-3B model. This suggests that scaling the capacity of the model is one of the most important factors, and should be considered as a top priority in future multi-modal dual encoder research.

\custompara{Results on Human Set and Human Performance.} Table~\ref{tab:main_results_test} shows that the results on the test set and human set are generally aligned with observations on the validation set. We conduct a study to estimate the human performance on \datasetname, via requesting 3 dedicated human annotators to answer 100 examples (sampled from human 
evaluation set, answers are non-overlapping).
We allow the annotators to use search engines (\eg, Google Image Search, Wikipedia Search, etc.)\footnote{Even with search engines, each annotator has used 254 seconds to complete one example.}, as long as the annotators can provide a valid Wikipedia entity name as the answer. As a result of this study, human achieves 77.7\% harmonic mean accuracy, which is significantly higher than the best comparison systems shown in Table~\ref{tab:main_results_test}.

%% file: tables_and_figures/exp_main_val_table.tex
\begin{table*}
    \centering
	\tabcolsep 5pt
    {
	    \begin{tabular}{lc@{\quad}ccc@{\quad\quad}ccc@{\quad\quad}c}
	      \toprule
	      & & \multicolumn{3}{c}{Entity Split$_{(\texttt{Dev})}$} & \multicolumn{3}{c}{Query Split$_{(\texttt{Dev})}$} & Overall$_{(\texttt{Dev})}$ \\
	      \addlinespace
	      & \# Params & {\seen} & {\unseen} & \textsc{hm} & {\seen} & {\unseen} & \textsc{hm} & \textsc{hm} \\ \midrule
	      \multicolumn{3}{l}{\textbf{Dual Encoders:}} \\
	      \xspace\BMarker{gray}\xspace$\text{CLIP}_{\xspace\texttt{ViTL14}}$ & \texttt{0}.\texttt{42B} & \pz5.4 & \pz5.3 & \pz5.4 & \pz0.8 & \pz1.4 & \pz1.0 & \pz1.7 \\
	      \xspace\BMarker{codegreen}\xspace$\text{CLIP Fusion}_{\xspace\texttt{ViTL14}}$ & \texttt{0}.\texttt{88B} &  32.7 & \pz4.3 & \pz7.7 & 33.4 & \pz 2.2 & \pz 4.2 & \pz5.4 \\
	      \xspace\BMarker{citecolor}\xspace$\text{CLIP2CLIP}_{\xspace\texttt{ViTL14}}$   & \texttt{0}.\texttt{86B} & 12.6 & 10.1 & 11.2 & \pz4.1 & \pz2.1 & \pz2.8 & \pz4.4 \\
	      \multicolumn{3}{l}{\textbf{Encoder Decoder:}} \\
	      \xspace\EMarker{codepurple}\xspace$\text{{PaLI-3B}}$ &  \pz\pz\pz\texttt{3B} & 21.6	&6.6 &10.1& 33.2&	14.7&		20.4&		13.5\\
	      \xspace\EMarker{cerise}\xspace$\text{{PaLI-17B}}$ & \pz\pz\texttt{17B} & 30.6	& 12.4 &		17.6 &  44.2	& 22.4 & 29.8	& 22.1 \\
	     \bottomrule
	    \end{tabular}
	}
	\caption{
	    Comparison between the fine-tuned models on the \datasetname \textbf{validation} set. %
    }
   
    \label{tab:main_results_validation}
\end{table*}

%% file: tables_and_figures/exp_main_test_table.tex
\begin{table*}
    \centering
        	\tabcolsep 6pt
    	    \begin{tabular}{l@{\;}ccccccccc}
    	     \toprule
    	      & & \multicolumn{2}{@{\;}c@{\;}}{Entity Split$_{(\texttt{Test})}$} & \multicolumn{2}{@{\;}c@{\;}}{Query Split$_{(\texttt{Test})}$} & Overall$_{(\texttt{Test})}$ & \multicolumn{3}{c}{Human Eval}\\ \addlinespace
    	      & \# Params & {\seen} & {\unseen} & {\seen} & {\unseen} & \textsc{hm} & {\seen} & {\unseen} & \textsc{hm} \\
    	     \midrule
    	     \multicolumn{3}{l}{\textbf{Dual Encoders:}} \\
             \xspace\BMarker{gray}\xspace$\text{CLIP}_\texttt{ViTL14}$ & \texttt{0}.\texttt{42B} & \pz5.6 & \pz4.9 & \pz1.3 & \pz2.0 & \pz2.4 & \pz4.6 & \pz6.0 & \pz5.2 \\
             \xspace\BMarker{codegreen}\xspace$\text{CLIP Fusion}_\texttt{ViTL14}$  & \texttt{0}.\texttt{88B} & 33.6 & \pz4.8 & 25.8 & \pz1.4 & \pz4.1 & 18.0 & \pz2.9 & \pz5.0\\
    	     \xspace\BMarker{citecolor}\xspace$\text{CLIP2CLIP}_\texttt{ViTL14}$  & \texttt{0}.\texttt{86B} & 12.6 & 10.5 & \pz3.8 & \pz3.2 & \pz5.3 & 14.0 & 11.1 & 12.4\\
    	     \multicolumn{3}{l}{\textbf{Encoder Decoder:}} \\
	         \xspace\EMarker{codepurple}\xspace$\text{{PaLI-3B}}$ & \pz\pz\pz\texttt{3B} & 19.1	&6.0& 27.4&	12.0& 11.8& 30.5&	15.8& 20.8 \\
	         \xspace\EMarker{cerise}\xspace$\text{{PaLI-17B}}$ & \pz\pz\texttt{17B} & 28.3 & 11.2 & 36.2 & 21.7 & 20.2 & 40.3 & 26.0 & 31.6 \\
    	     \midrule
    	     \textbf{Human+Search \protect\footnotemark} \cellcolor{lightgray} & \cellcolor{lightgray} - & \cellcolor{lightgray} -
    	     & \cellcolor{lightgray} - & \cellcolor{lightgray} -
    	     & \cellcolor{lightgray} - & \cellcolor{lightgray} - & 76.1\cellcolor{lightgray} & 79.3 \cellcolor{lightgray} & 77.7 
    	     \cellcolor{lightgray} \\
    	     \bottomrule
    	    \end{tabular}
	\caption{
	    Results of methods on the \datasetname \textbf{test} set and \textbf{human evaluation} set. 
	    Human+Search represents human performances with information retrieval tools such as search engines and others, on a random subset of $\text{\datasetname}_{\texttt{Human\_Eval}}$. 
    }
    \label{tab:main_results_test}
\end{table*}

\footnotetext{The human study is done on a random sampling of 100 examples.}

%% file: sections/analysis.tex
In this section, we perform empirical studies to analyze the pre-trained CLIP2CLIP and PaLI models, and conduct a detailed analysis of these two models' common errors.

\input{tables_and_figures/exp_ablation_finetune}

\custompara{Does fine-tuning always help generalization?} Figure~\ref{fig:finetune} presents the validation scores of the PaLI model (left) and the CLIP2CLIP model (right), during fine-tuning on \datasetname's training split. It shows that a longer training schedule does not lead to better generalization performance, particularly when evaluated on the \unseen entities. Because of this, we employ the early stopping strategy for model selection, and pick the model with the best harmonic mean combined score on the validation set. However, due to this early stopping strategy, both fine-tuned models are not utilizing 100\% of the examples in  \name's training data because their \unseen performance starts to degenerate within one epoch. This has indicated that more advanced fine-tuning strategies that use better regularization techniques to encourage generalization across Wikipedia entities, could be a promising research to explore in the future. 

\input{tables_and_figures/exp_ablation_kb_sizes}

\custompara{How would the number of entities in KB influence the model's prediction?} Figure~\ref{fig:kbsizes} presents the accuracy of CLIP2CLIP, as a function of the \# of total candidates to retrieve from. Here, we compute the accuracy by sub-sampling the negative candidates from KB to different sizes. We observe that when the retrieval candidate entities are only the positive entities (with the \# of candidates being 20K), the performance of the CLIP2CLIP model is significantly higher than the open-domain setting (with 6M entities in total). 
Beyond this, as the KB size increases, model accuracy decreases. Concretely, it shows an approximately linear decline along the log-scale x-axis in Figure~\ref{fig:kbsizes}. This indicates that as the KB size increases, the models' accuracy first drops significantly and then follows with a gradual decline. On the other hand, PaLI's performance is generally more steady as the size of KB grows, potentially because its prediction has already matched up entity names inside KB, so narrowing down the set of candidates does not help the BM25 post-processing. One potential direction is to employ constrained decoding for the PaLI-based model, which we leave for future works.

\custompara{How would models perform on head vs. tail entities?} We evaluate the visual entity recognition performances of CLIP2CLIP and PaLI, on entities of different popularity. Particularly, Figure~\ref{fig:head2tail} presents a histogram according to models' performance on the entity that has different average monthly Wikipedia page views in 2022~\cite{mallen2022not}. From the comparison, we can see that PaLI is significantly more accurate compared to CLIP2CLIP, on the head entities (that have more than 5K monthly page views). However, we observe that CLIP2CLIP can perform on par or even outperform PaLI on tail-ish entities (that have less than 2.5K monthly views). This suggests that the retrieval-based visual entity recognition model has its own advantages, in recognizing the difficult and tail entities. Meanwhile, this result also provides a hint that potentially a frequency calibrated evaluation should be developed to reward models more with strong recognition capability on the tail entities.

\input{tables_and_figures/exp_ablation_head2tail}

\custompara{Error analysis} To better understand the errors that CLIP2CLIP and PaLI models are making, we sampled a random 100 examples on the human evaluation set, and manually categorize and analyze the errors that PaLI and CLIP2CLIP are making. Particularly, we categorize the errors of the pre-trained models into four categories: (a) erroneous but relevant prediction, on concepts of the same granularity; (b) errors due to predicting very generic concepts; (c) errors due to misunderstanding the intent behind the query. (d) other miscellaneous errors. Note that errors type (d) are mostly mistakes that are unrelated and not easily interpretable. The results are shown in Table~\ref{tab:error_analysis}. Figure~\ref{fig:error_analysis} has provided some concrete examples of the above types of mistakes made by CLIP2CLIP and PaLI. Interestingly, it shows that the two models, \ie, CLIP2CLIP and PaLI, are making very different types of errors in their predictions. Particularly, CLIP based model is good at capturing the right granularity of the entity, but often fails to understand the true intent of the text query. For instance, Figure~\ref{fig:error_analysis} (c) shows that CLIP2CLIP ignores the text query and starts to predict the name of the barrel racer. In contrast, PaLI is good at following the text query, but can usually predict generic concepts when it does not know the answer confidently (see Figure~\ref{fig:error_analysis} (b)). 

\input{tables_and_figures/exp_ablation_error_type}
\input{tables_and_figures/exp_ablation_error_type_example}

%% file: tables_and_figures/exp_ablation_finetune.tex
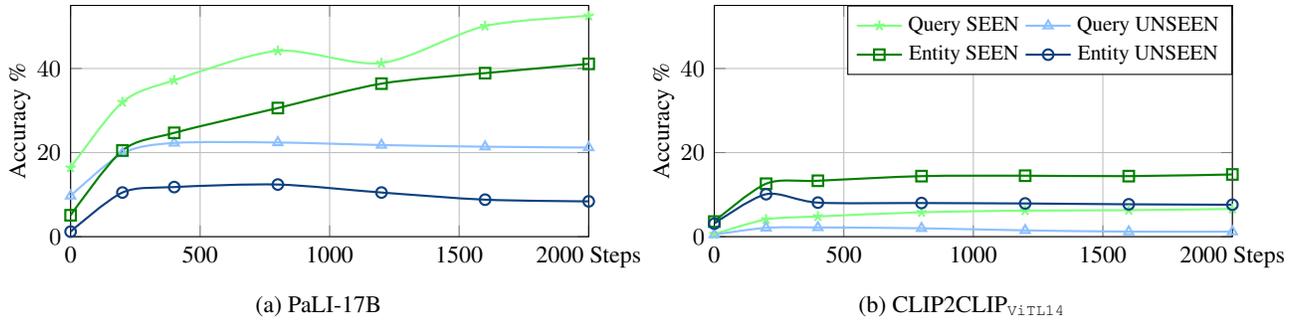
\begin{figure*}[htp]
    \centering
    \begin{subfigure}[]{.485\textwidth}
       \begin{tikzpicture}
            \begin{axis}[
            	width=1\textwidth,
        	    height=0.55\columnwidth,
        	    legend style={at={(1, 1)},anchor=north east,font=\footnotesize},
        		font=\small,
        		xlabel near ticks,
        		ylabel near ticks,
        	    xmin=0,xmax=2000,
           		ymin=0, ymax=55,
           	 	ymajorgrids=true,
            	xmajorgrids=true,
                xtick={0, 500, 1000, 1500, 2000},
                xticklabels={0, 500, 1000, 1500, 2000 Steps},
                ylabel=Accuracy \%,
            	ylabel style={yshift=-1ex,},
            	legend columns=2]
            
                \addplot[smooth,mark=star,green!50,thick] plot coordinates {
                    (0, 16.4)
                    (200,32.0)
                    (400,37.2)
                    (800,44.2)
                    (1200,41.3)
                    (1600,50.1)
                    (2000,52.5)
                };
                
                \addplot[smooth,mark=triangle,citecolor!50,thick] plot coordinates {
                    (0, 9.7)
                    (200,19.8)
                    (400,22.3)
                    (800,22.4)
                    (1200,21.8)
                    (1600,21.4)
                    (2000,21.2)
                };
                \addplot[smooth,mark=square,green!50!black,thick] plot coordinates {
                    (0, 5.1)
                    (200,20.5)
                    (400,24.7)
                    (800,30.6)
                    (1200,36.4)
                    (1600,38.9)
                    (2000,41.1)
                };
                
                \addplot[smooth,mark=o,citecolor!50!black,thick] plot coordinates {
                    (0, 1.2)
                    (200,10.5)
                    (400,11.8)
                    (800,12.4)
                    (1200,10.5)
                    (1600,8.8)
                    (2000,8.4)
                };
            \end{axis}
        \end{tikzpicture}
        \caption{PaLI-17B}
    \end{subfigure}
    \begin{subfigure}[]{.485\textwidth}
       \begin{tikzpicture}
            \begin{axis}[
            	width=1\textwidth,
        	    height=0.55\columnwidth,
        	    legend style={at={(1, 1)},anchor=north east,font=\footnotesize},
        		font=\small,
        		xlabel near ticks,
        		ylabel near ticks,
        	    xmin=0,xmax=2000,
           		ymin=0, ymax=55,
           	 	ymajorgrids=true,
            	xmajorgrids=true,
                xtick={0, 500, 1000, 1500, 2000},
                xticklabels={0, 500, 1000, 1500, 2000 Steps},
                ylabel=Accuracy \%,
            	ylabel style={yshift=-1ex,},
            	legend columns=2]

                \addplot[smooth,mark=star,green!50,thick] plot coordinates {
                    (0, 0.6)
                    (200,4.1)
                    (400,4.8)
                    (800,5.8)
                    (1200,6.2)
                    (1600,6.3)
                    (2000,6.6)
                };
                \addlegendentry{Query SEEN}
                
                \addplot[smooth,mark=triangle,citecolor!50,thick] plot coordinates {
                    (0, 0.5)
                    (200,2.1)
                    (400,2.2)
                    (800,2.0)
                    (1200,1.5)
                    (1600,1.2)
                    (2000,1.2)
                };
                \addlegendentry{Query UNSEEN}
                \addplot[smooth,mark=square,green!50!black,thick] plot coordinates {
                    (0, 3.6)
                    (200,12.6)
                    (400,13.3)
                    (800,14.4)
                    (1200,14.5)
                    (1600,14.4)
                    (2000,14.8)
                };
                \addlegendentry{Entity SEEN}
                
                \addplot[smooth,mark=o,citecolor!50!black,thick] plot coordinates {
                    (0, 3.1)
                    (200,10.1)
                    (400,8.1)
                    (800,8.0)
                    (1200,7.9)
                    (1600,7.7)
                    (2000,7.6)
                };
                \addlegendentry{Entity UNSEEN}
            \end{axis}
        \end{tikzpicture}
        \caption{CLIP2CLIP$_\texttt{ViTL14}$}
    \end{subfigure}
    \caption{
        \textbf{Fine-tuning PaLI or CLIP2CLIP for large \# of steps} increases the \seen entity accuracy but hurts the \unseen entity accuracy.
    }
    \vspace{-2mm}
    \label{fig:finetune}
\end{figure*}

%% file: tables_and_figures/exp_ablation_kb_sizes.tex
\begin{figure*}
    \centering
    \begin{subfigure}[]{.485\textwidth}
        \begin{tikzpicture}
        \begin{semilogxaxis}[
            	width=1\columnwidth,
        	    height=0.55\columnwidth,
        	    legend style={at={(1, 1)},anchor=north east,font=\footnotesize},
        	    mark options={mark size=2},
        		font=\small,
        		log ticks with fixed point,
        		xlabel near ticks,
        		ylabel near ticks,
        	    xmin=0,xmax=650,
           		ymin=0, ymax=35,
           	 	ymajorgrids=true,
            	xmajorgrids=true,
            	xtick={2, 10, 100, 600},
                xticklabels={20K, 100K, 1M, 6M},
                ylabel=HM Accuracy \%,
            	ylabel style={yshift=-1ex,}]
         
            \addplot[smooth,mark=o,navyblue!25!black,thick] plot coordinates {
                (2, 30.9)
                (10,30.8)
                (100,30.4)
                (600,29.8)
            };
            
            \addplot[smooth,mark=square,red,thick] plot coordinates {
                (2, 9.21)
                (10,6.38)
                (100,4.23)
                (600,2.75)
            };
        \end{semilogxaxis}
        \end{tikzpicture}
        \caption{Query Split}
    \end{subfigure}
  \begin{subfigure}[]{.485\textwidth}
        \begin{tikzpicture}
        \begin{semilogxaxis}[
            	width=1\columnwidth,
        	    height=0.55\columnwidth,
        	    legend style={at={(1, 1)},anchor=north east,font=\footnotesize},
        	    mark options={mark size=2},
        		font=\small,
        		log ticks with fixed point,
        		xlabel near ticks,
        		ylabel near ticks,
        	    xmin=0,xmax=650,
           		ymin=0, ymax=35,
           	 	ymajorgrids=true,
            	xmajorgrids=true,
            	xtick={2, 10, 100, 600},
                xticklabels={20K, 100K, 1M, 6M},
                ylabel=HM Accuracy \%,
            	ylabel style={yshift=-1ex,}]
         
            \addplot[smooth,mark=o,navyblue!50!black,thick] plot coordinates {
                (2, 19.8)
                (10,19.1)
                (100,18.5)
                (600,17.6)
            };
            \addlegendentry{PaLI-17B}

            \addplot[smooth,mark=square,red,thick] plot coordinates {
                (2, 20.5)
                (10,17.31)
                (100,15.50)
                (600,11.23)
            };
            \addlegendentry{CLIP2CLIP}
            
        \end{semilogxaxis}
        \end{tikzpicture}
        \caption{Entity Split}
    \end{subfigure}
    \caption{
        \textbf{Impact of \# Wikipedia Candidates on PaLI and CLIP2CLIP.} Increasing the size of Wikipedia makes the tasks difficult.}
    \label{fig:kbsizes}
    \vspace{-10px}
\end{figure*}
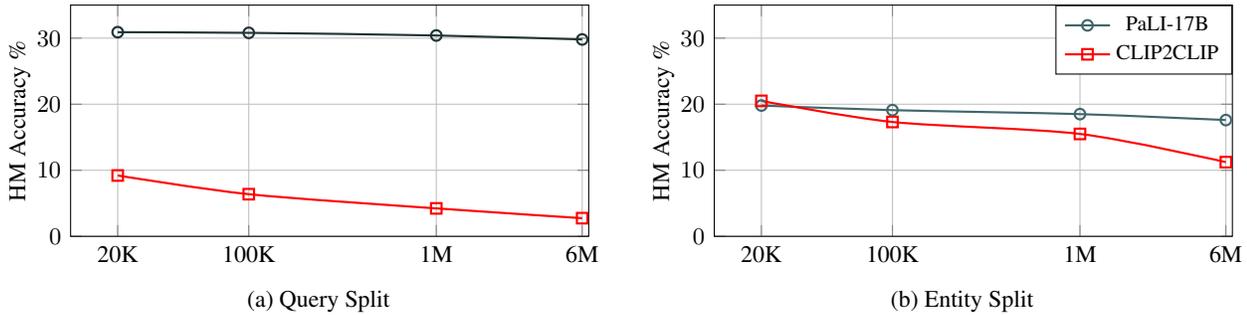

%% file: tables_and_figures/exp_ablation_head2tail.tex
\begin{figure}[t]
    \centering
    \begin{tikzpicture}
    \begin{axis}[
        width=1\columnwidth,
        height=0.55\columnwidth,
        ybar,
        ymin=0,
        ymax=55,
        bar width=2.5mm,
		xlabel near ticks,
		ylabel near ticks,
        xtick=data,
        font=\small,
        xticklabels={100, 500, 1K, 2.5K, 5K, 10K, 25K, 50K, 100K},
        xlabel={Average Monthly View of Entity},
        ylabel={Accuracy \%},
        legend columns=2,
        legend style={font=\small},
    ]
    \addplot[fill=red!30] coordinates {
        (0, 1.87)
        (1, 4.73)
        (2, 8.01)
        (3, 10.35)
        (4, 12.31)
        (5, 14.40)
        (6, 11.12)
        (7, 9.66)
        (8, 5.41)
    };
    \addplot[fill=citecolor!50] coordinates {
        (0, 0.65)
        (1, 1.7)
        (2, 4.76)
        (3, 11.45)
        (4, 17.14)
        (5, 27.24)
        (6, 32.28)
        (7, 38.89)
        (8, 30.87)
    };
    \legend{CLIP2CLIP, PaLI-17B}
    \end{axis}
    \end{tikzpicture}
    \caption{
    \textbf{Comparison of Performances on Head vs. Tail Entities (results on Validation set).} PaLI wins over CLIP2CLIP on popular (\ie, high monthly page view) Wikipedia entities, but loses on rare (\ie, low monthly page view) Wikipedia entities. 
    }
    \label{fig:head2tail}
\end{figure}
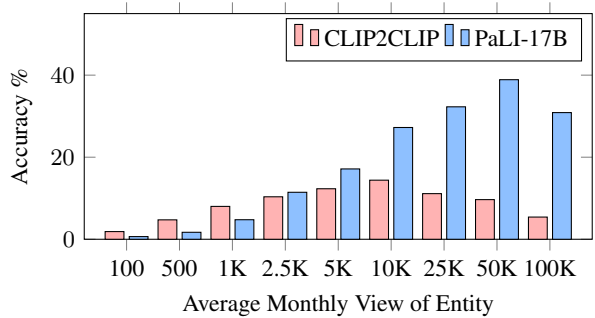

%% file: tables_and_figures/exp_ablation_error_type.tex
\begin{table}[tb]
\begin{center}
\begin{small}
\tabcolsep 4pt
\begin{sc}
\begin{tabular}{lrr}
\toprule
 & PaLI-17B & CLIP2CLIP  \\
\midrule
Correct & 29\% & 15\% \\
\rowcolor{lightgray} In-correct & 71\% & 85\% \\
{\xspace$\rightarrow$} (a) Wrong But Relevant & 23\% & 27\% \\ 
{\xspace$\rightarrow$} (b) Too Generic & 15\% & 1\%\\
{\xspace$\rightarrow$} (c) Misunderstand Query & 7\% & 37\%  \\
{\xspace$\rightarrow$} (d) Miscellaneous & 24\% & 20\% \\
\bottomrule
\end{tabular}
\end{sc}
\end{small}
\end{center}
\vskip -0.1in
\caption{Error type distribution for difference models. PaLI predicts more answers with less granularity (less granularity), while most of the CLIP errors are due to not understanding the questions.}
\label{tab:error_analysis}
\end{table}

%% file: tables_and_figures/exp_ablation_error_type_example.tex
\begin{table*}[t!]
    \begin{center}
    \tabcolsep 5pt
    {
    \begin{tabular}{@{\;\;}l@{\;\;}c c c@{\;\;}}
            \toprule
            Error Type & (a) Wrong but Relevant &  (b) Too Generic & (c) Misunderstand Query\\
            \midrule
            Input Query & \makecell{\it What is the name of the model \\ of this aircraft?} & \it \makecell{What is the species of \\ this animal?} & \it \makecell{What sports event is displayed \\ in the picture?} \\
            Input Image & \includegraphics[width=40mm,height=30mm]{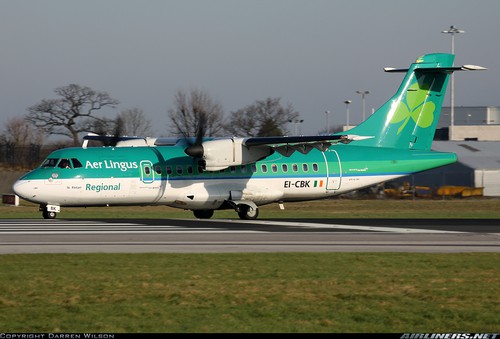}
            & \includegraphics[width=35mm,height=30mm]{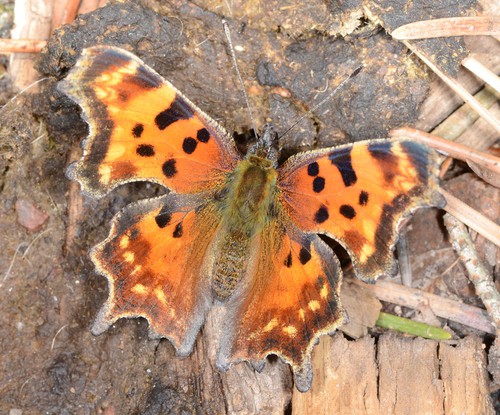} & \includegraphics[width=40mm,height=30mm]{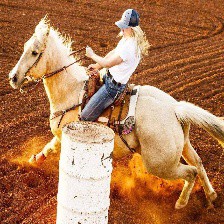} \\
            \midrule
            PaLI-17B: & \begin{tabular}{|l|} \hline
                    \textcolor{cerise}{\snlp{WikiID: Q589498}} \\
                    \textcolor{cerise}{\snlp{Name:} \textit{BAe 146}} \\
                    {\includegraphics[width=40mm,height=30mm]{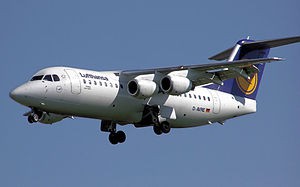}} \\ \hline
                    \end{tabular}
                  & \begin{tabular}{|l|} \hline
                    \textcolor{cerise}{\snlp{WikiID: Q255496}} \\
                    \textcolor{cerise}{\snlp{Name:} \textit{Butterfly}} \\
                    {\includegraphics[width=35mm,height=30mm]{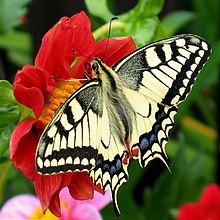}} \\ \hline
                    \end{tabular}
                  & \begin{tabular}{|l|} \hline
                    \textcolor{darkgreen}{\snlp{WikiID: Q2529836}} \\
                    \textcolor{darkgreen}{\snlp{Name:} \textit{Barrel racing}} \\
                    {\includegraphics[width=35mm,height=30mm]{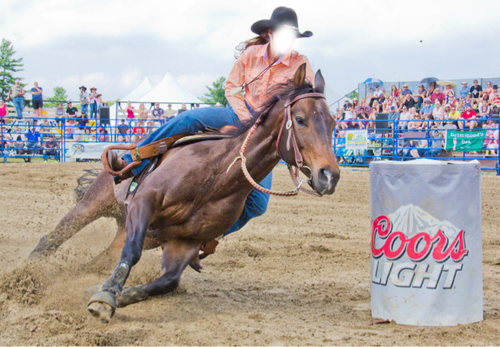}} \\ \hline
                    \end{tabular} \\ \addlinespace
            CLIP2CLIP: & \begin{tabular}{|l|} \hline
                    \textcolor{cerise}{\snlp{WikiID: Q937949}} \\
                    \textcolor{cerise}{\snlp{Name:} \textit{Dornier 328}} \\
                    {\includegraphics[width=40mm,height=30mm]{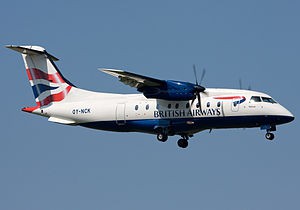}} \\ \hline
                    \end{tabular}
                  & \begin{tabular}{|l|} \hline
                    \textcolor{cerise}{\snlp{WikiID: Q13510645}} \\
                    \textcolor{cerise}{\snlp{Name:} \textit{Proteuxoa comma}} \\
                    {\includegraphics[width=35mm,height=30mm]{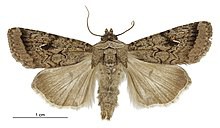}} \\ \hline
                    \end{tabular}
                  & \begin{tabular}{|l|} \hline
                    \textcolor{cerise}{\snlp{WikiID: Q****4678}} \\
                    \textcolor{cerise}{\snlp{Name:} \textit{E. W. (barrel racer)}$\dagger$} \\
                    {\includegraphics[width=40mm,height=30mm]{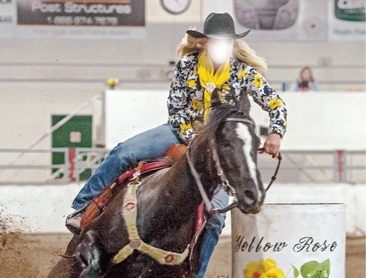}} \\ \hline
                    \end{tabular} \\ \addlinespace \midrule
            Ground-Truth: & \begin{tabular}{|l|} \hline
                    {\snlp{WikiID: Q218637}} \\
                    {\snlp{Name:} \textit{ATR 42}} \\
                    {\includegraphics[width=40mm,height=30mm]{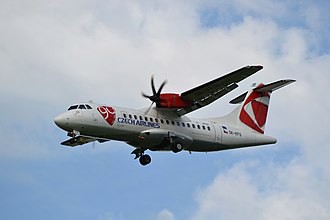}} \\ \hline
                    \end{tabular}
                  & \begin{tabular}{|l|} \hline
                    {\snlp{WikiID: Q592001}} \\
                    {\snlp{Name:} \textit{Hoary comma}} \\
                    {\includegraphics[width=35mm,height=30mm]{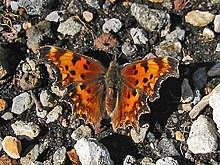}} \\ \hline
                    \end{tabular}
                  & \begin{tabular}{|l|} \hline
                    {\snlp{WikiID: Q2529836}} \\
                    {\snlp{Name:} \textit{Barrel racing}} \\
                    {\includegraphics[width=35mm,height=30mm]{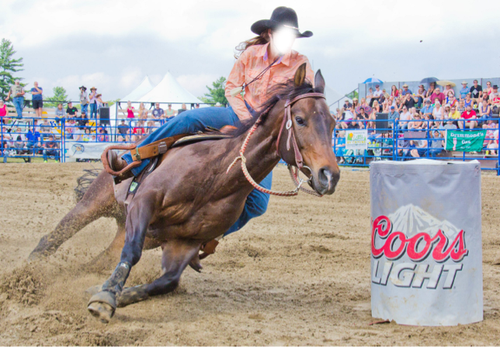}} \\ \hline
                    \end{tabular} \\ \bottomrule
    \end{tabular}
    }
    \caption{
        \textbf{Visualization of mistakes made by the CLIP2CLIP and PaLI-17B Model.} We visualize the Wikipedia infobox images for each of model's predictions, to provide more context about the visual similarity between the prediction/ground-truth and the input image. Correct predictions are marked as \textcolor{darkgreen}{green}, whereas incorrect predictions are marked as \textcolor{cerise}{red}. ($\dagger$: Since no infobox image is available for this Wikipedia entity, a face-anonymized Web image of the entity is visualized for reference.)
    }
    \vspace{-5mm}
    \label{fig:error_analysis}
    \end{center}
\end{table*}

%% file: sections/related.tex
\custompara{Learning to Recognize \unseen Categories} There has been a significant amount of prior work~\cite{lampert2014attribute,vinyals2016matching,liu2019large} focusing on the generalization situation where information of novel categories are presented at test time. Zero-shot learning (ZSL) is one of such attempts that tackles learning new categories with zero images for training. To achieve such transfer, ZSL methods typically rely generating \unseen image classifiers based on corresponding semantic representations, in the format of manually labeled attributes~\cite{lampert2014attribute}, unsupervised learned word vectors~\cite{changpinyo2016synthesized}, or pre-trained sentence embeddings~\cite{kil2021revisiting,radford2021learning}. Few-shot learning (FSL)~\cite{vinyals2016matching} proposes a more realistic setup, where learners have access to a limited number of visual exemplars during the model deployment. With this goal, FSL methods aim to extract the inductive bias of learning from the \seen classes, such that the model can leverage it in learning the \unseen classes, to avoid severe over-fitting. Particularly, prior works either use adapted non-parametric classifiers~\cite{snell2017prototypical,Rusu2018Meta,ye2020few}, or meta-optimized linear classifiers~\cite{finn2017model,raghu2019rapid} to incorporate the few-shot \unseen support examples. Comparing to them, our proposed task exposes different challenges as we ask the model to make the best use of open-world Web knowledge (\ie, Wikipedia pages with images \& figures), which contains textual semantic information and visual appearance of the entities in the open world.

\custompara{Vision and Language + Knowledge} There have been efforts in combining knowledge into vision and language tasks, such as Visual QA~\cite{shah2019kvqa,marino2019ok,chang2022webqa,chen2021zero} and entity-focused image captioning~\cite{liu2020visual,biten2019good}. Among them, knowledge-based VQA is most related to \name, but also differs in many aspects. Specifically, \cite{chang2022webqa} presents a text QA dataset that requires understanding multi-modal knowledge in a KB. \cite{shah2019kvqa} propose to perform knowledge-based question answer tasks, centered around questions that resolve relational query over public Figures. Meanwhile, \cite{marino2019ok} propose to answer questions where the answer is outside of the image context, to assess model's capability in understanding real-world knowledge More recently, \cite{chen2021zero} studies the zero-shot visual QA setting where some answers (out of a total of 500 frequent answers of general concepts) are unseen during the training, where a KB is supplied to assist the model in answering unseen answers. Comparing to them, \name steps back to the more fundamental problem of establishing the link between visual content and entity in the KB, but at a larger scale and broader coverage. We believe that stronger models developed on \name would benefit such knowledge-intensive visual QA tasks. 

\custompara{Entity Linking} Entity linking (EL) is the task of grounding entity mentions in the text by linking them to entries in a given
knowledge base. 
Supervised EL~\cite{milne2008learning} has demonstrated its strong performance when all entities are in-distribution during the evaluation. Because KB is updating all the time, recent works~\cite{logeswaran2019zero,botha2020entity,zhang2021entqa,de2020autoregressive,de2022multilingual} focus on a more realistic setting where entity linking needs to be achieved in the zero-shot, with a large portion of entities (to be evaluated) completely unseen during the training. \name is a visual analog of zero-shot EL, and targets at developing generalizable models that recognize entities unseen in the training. Among all EL literature, visually assisted EL~\cite{zheng2022visual} is most relevant to this work, whose goal is to use the associated image of text to improve the precision of text EL. \name is different as its text queries do not mention the name of the entities, which put visual understanding and reasoning into the central position. 

%% file: sections/discussion.tex
In this paper, we have introduced \name, a task that aims to unambiguously link visual content to the corresponding entities in a web-scale knowledge base (\ie, Wikipedia), covering a total of more than 6 millions of entities. To facilitate the evaluation of \name, we created the \datasetname dataset, via combining and re-annotating 14 existing visual recognition, retrieval, and visual QA datasets, and linked over 20K labels to the Wikipedia entities. With \datasetname, we evaluate state-of-the-art multi-modal pre-trained models, \ie, the CLIP~\cite{radford2021learning}-based entity retrieval models and the PaLI~\cite{chen2022pali}-based entity generation model, via fine-tuning them for the \name task, to examine their capability on recognizing open-domain visual concepts. As a result, PaLI models have presented significantly stronger performances than the CLIP variants, even on unseen visual entities during the fine-tuning. Meanwhile, although the CLIP-based entity retrieval model is overall weaker, it shows advantages in recognizing the tail visual entities. 

One additional nice property of \datasetname is its strong extensibility. As a result of grounding of all recognition labels to Wikipedia entities, we as a community can keep growing the member recognition datasets of \datasetname, by adding positive instances to Wikipedia entities that do not have examples by far. Moreover, successful \name models can generalize to recognize emerging entities (\eg, \nlp{iPhone 14 Pro}), as long as the corresponding Wikipedia page is created. In summary, we hope \name will drive future research on knowledge-infused multimodal representation learning via visual entity recognition.

%% file: sections/appendix.tex
\section{Dataset Construction, Annotation, and Additional Statistics}

In this section, we describes the complete details on data collection, curation, entity linking, and show additional statistics of the processed dataset (\myref{appendix:data_collection}). Then we also discuss how we train annotators to annotate our task, and provide the concrete annotation interface(\myref{appendix:human_annotation}).

\subsection{Data Collection \& Pre-processing}
\label{appendix:data_collection}

\custompara{Data Filtering} 
Some of our member datasets have been reported to include non-imageable classes, classes with undesired social bias~\cite{yang2020towards}, or non-entity classes (\eg, numbers). Therefore, we apply a filtering process to compose our dataset, based on the individual condition of each source dataset. Overall, to create the Entity split, we first apply a general safety filter~\cite{yang2020towards} to remove non-imageable labels, non-entity labels, and labels with social bias.
To create the Query split, we employed three expert annotators to write heuristic policies to filter each VQA dataset, and ensure our task is focusing on entity related questions. Concretely, questions related to counting, verification, or querying non-entity attributes (\eg, dates), are removed. Then we apply the same safety filter.

\custompara{Linking labels to Wikipedia Entities} Based on the filtered data, we developed a two-staged entity linking strategy to connect the label text to Wikipedia entities, on both Entity and Query splits.
First, we obtain exact match based entity candidates by querying the Wikipedia search API (with the auto-suggestion disabled) with the raw label text.
We reject candidates whose landing pages are identified as disambiguation pages. The Wikipedia API\footnote{ \url{https://www.mediawiki.org/wiki/API}} automatically redirects queries (in our case, labels) matching entity aliases to their canonical form.
For the labels which do not have an exact match in Wikipedia, we use a state-of-the-art text-based entity linker (\ie, GENRE~\cite{decao2021autoregressive}) to obtain top candidate Wikipedia entity names.
Finally, we link the label to the top ranked entity whose landing page is not a disambiguation page. 

\custompara{Preparing Multi-Modal Knowledge} Using the entity linking process described earlier, we successfully connect a total of 24,895 class labels in \datasetname to corresponding Wikipedia entities. Overall, our dataset contains 20,801 unique entities.
For the Entity split data, we generate a synthetic text query based on the super-category information of the label (either provided by source dataset or mined from Wikidata\footnote{Available at \url{https://www.wikidata.org/wiki/Wikidata}}), using templated language. For example, iNaturalist has provided detailed supercategory annotation on each class, such as \nlp{Plantae}, \nlp{Reptilia}, \etc. For dataset that do not provide this information, we use the super-category mined from Wikidata, which is publiclly crowd sourced and maintained. As a result, our templated query generator produces the query \nlp{``what is the species of the plant in this image?''} for the entity \texttt{``Eryngium alpinum''}, whose super-category is \nlp{Plantae}. Due to space limit, we provide more explanation in Appendix. For all Wikipedia entities, we use the corresponding {Wikipedia page} and its associating {multi-media content} (\eg, information box images, \etc) as the source of {\em multi-modal knowledge} about entities. 

\input{tables_and_figures/appendix/dataset_unique_entities_appendix}

\custompara{Statistics on Entities} Specifically, Figure~\ref{fig:unique_entity_distribution} shows the number of unique entities in both the Entity and Query splits, where we compare the total number of entities in each source dataset against its original population (after applied safety filter). 
Note that for the Google Landmarks v2 (Gldv2) dataset, we employed the cleaned data split from~\cite{yokoo2020two}, where the total number of unique entities is significantly reduced. Because Gldv2 is automatically generated and has reported to contain noises particularly with tail entities~\cite{yokoo2020two}, we removed entities with less than 50 instances for a improved precision (further reduces the \# of entities in Gldv2 to $\sim$6k). 

\custompara{Entity Super-Categories} To give more details for the Figure~\ref{fig:dataset_statistics} in the main text, we further present full super-category grouping information in Figure~\ref{fig:dataset_stats_full}. As aforementioned, we have combined entities that belongs to general groups (\eg, ``object'', ``item'' groups) or unpopular groups (\eg, groups with less than 5 entities) into the ``others'' group. We also merged some sub-categories into super-categories, \eg, ``location''+``park''+``lake''+``river''+``mountain''$\rightarrow$``location'', ``building''+``bridge''$\rightarrow$``building''.

\input{tables_and_figures/appendix/dataset_stats_appendix}

\subsection{Human Annotation Procedure \& Interface}
In order to verify the quality of \datasetname and to provide a human verified test set to evaluate on, we conduct human annotation on a subset of test set. The annotators are asked to correct the errors in the  <image, query, answer> triplets. The details are as follows.
\label{appendix:human_annotation}

\input{tables_and_figures/appendix/mixmore_annotation_interface}

\paragraph{Annotation interface} Figure~\ref{fig:annotation_interface} illustrates the annotation interface. The left side of Figure~\ref{fig:annotation_interface} are the input to the annotators which includes the original question, image and the answer (together with the wikipedia hyperlink). The annotators are asked to complete the following questions:
\begin{enumerate}[leftmargin=*,topsep=5pt,itemsep=1pt]
    \item \textit{Does the Wikipedia represent the correct meaning of the answer?  Provide the Wikipedia link if not. }  
    
    This question requires the annotators to correct the entity linking errors.
    The annotators use Google search to find the most suitable Wikipedia link if the provided one is not adequate. In our dataset, 8.4\% of the entity links are reported wrong by more than 2 annotators, which are manually corrected later.
    
    \item \textit{Is the Wikipedia answer physically present in the image.}
    
    This question is mainly aimed at filtering out the OCR examples which are out of our scope. One example is that the image about a wall painted with the word ``love" and the linked entity is the ``love" Wikipedia. In our dataset, 10.3\% of the answers are reported not physically present in the image by more than 2 annotators, which are discarded from the human evaluation set.
    
    \item \textit{Rewrite the question so that no other object can be the answer.} 
    
    The annotators will rewrite the question is the answer is wrong or ambiguous. Annotators will make sure that the question can not be answered without the image and that the answers can not be included in the rewritten questions. In our annotation, 99.9\% of the questions are being rewritten.
    
\end{enumerate}

\paragraph{Instruction and Training}
We carefully design the training procedure to improve the annotation quality. We first conduct a ``Self-study session'' where the annotators will read the instructions and annotate a few toy examples. Then we conduct a ``In-person tutorial'' where we have an online video session in which we walk annotators through the full version of the instructions and discuss mistakes made in the self-study annotations. Finally we conduct a ``Test exam'' and the qualified annotators are accepted. In total, 30 annotators went through our training procedure and all of them were eventually accepted to work full-time on the main task.

\paragraph{Quality control}
We have a three way annotations where each examples are annotated by three annotators. We were giving regular feedback on the questions the annotators may have during the annotation and pointed out mistakes identified in annotators' past answers.

On average, it took annotators 4.6 minutes to answer each question with the time consumption slightly decreasing as annotators get familiar with the task. The compensation rate for the task was set to be \$17.8/hour which is higher than the minimum hourly wage in the US.

We filtered out all the examples where the wikipedia links are marked as wrong or  the Wikipedia answers are marked as  ``Not physically present in the image". 

\section{Implementation Details of the baseline systems}
\label{appendix:implementation}

In this section, we provide implementation details on the CLIP variants and PaLI model for the \name task.

\subsection{CLIP Fusion Model}
As aforementioned, we implemented this multi-modal dual encoder via taking pre-trained CLIP image and text encoders as featurizers. The CLIP model is based on a ViT-Large, with a total of over 400\texttt{M} parameters, pre-trained on a 400\texttt{M} prviate image-text dataset collected by OpenAI. Based on this model, we build two 2-layer Transformer models, on top of two CLIP models as the left and right encoder, for encoding the query representation and the entity representation, respectively. The 2-layer Transformers follows the same architecture as T5 Transformer~\cite{radford2021learning}, but with 2 layers, 12 attention heads, with each attention head of 64 dimensions, and the embedding size of 768. We then fine-tune this composed model on the \datasetname's training data, using a in-batch contrastive learning objective~\cite{radford2021learning}, with a batch size of 4,096. We optimize the model for 10K steps in the fine-tuning stage, with Adafactor optimizer~\cite{shazeer2018adafactor} and a initial learning rate of 0.001. There are 1k steps for the warmup, followed by a square root LR decay schedule with final learning rate of 1e-6.

\subsection{CLIP2CLIP Model}
Different from CLIP Fusion, CLIP2CLIP is a model that adds minimum new parameters to the pre-trained CLIP encoders. Same as other models, we initialize both the query encoder and the target encoder separately with the pre-trained CLIP model. Specifically, we use the pre-trained CLIP encoders for both left and right encoders, to encode the image and text modality for both the query representation and the entity representation. We then compute the four dot product similarity scores on the $\texttt{<}$input image, target text$\texttt{>}$, $\texttt{<}$input text, target image$\texttt{>}$, $\texttt{<}$input image, target image$\texttt{>}$, and $\texttt{<}$input text, target text$\texttt{>}$ pairs, which is then combined via a learnable similarity weights into one logit score. The make sure that the learnable similarity weights is initialized properly, we perform a grid search to find a roughly good similarity weights for the CLIP2CLIP model (using \datasetname's training data). Then we took this similarity weights to initialize the CLIP2CLIP model and fine-tune all parameters on \datasetname's training set, under the same contrastive learning objective. Different from other models, given that this model has most of its parameters pre-trained, we realized that it works the best to early stop the model. As a result, we only fine-tune this model for 2k steps, with an initial learning rate of 1e-4, and a square root LR decay schedule with final learning rate of 1e-6. 

\subsection{PaLI Model}

As aforementioned, we have evaluated two variants of PaLI models, the model with 3B total parameters (\ie, PaLI-3B) and the model with 17B parameters (\ie, PaLI-17B).
The PaLI-17B model reuses 13B parameter from the mT5-XXL~\cite{xue2020mt5} and 4B parameters from the ViT-e~\cite{zhai2022scaling}, which were pre-trained Web Text and JFT-3B datasets, and then jointly trained on the WebLI~\cite{chen2022pali} dataset with 10\texttt{B} image and text pairs, under a variety of pre-training objectives, including object recognition, split captioning, visual question answering, etc. Similarly, the PaLI-3B model reuses 1B parameters from mT5-Large~\cite{xue2020mt5}, and 1.8B parameters from the  ViT-G~\cite{zhai2022scaling}, under the same pre-training recipe. To fine-tune PaLI on our dataset, we finetue the pre-trained PaLI model using its Visual QA interface, and inject the \name text queries into the PaLI's VQA prompt. As a concrete example, we convert a original query of \nlp{what species is the animal in the image?} into the format of \nlp{Answer in en: what species is the animal in the image?}, as input to the PaLI model. The objective of fine-tuning process is then to maximize the likelihood of answer generation, same as its standard VQA fine-tuning practices. Similarly, we employ the Adafactor optimizer to optimize the fine-tuning, with a total of 2K fine-tuning steps, with a warmup of 1K steps and linear LR decay schedule.

%% file: tables_and_figures/appendix/dataset_unique_entities_appendix.tex
\begin{figure*}[htp]
    \centering
    \begin{subfigure}[]{.545\textwidth}
        \includegraphics[width=\textwidth]{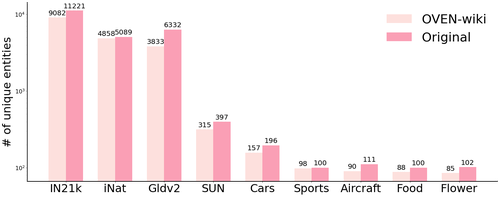}
    \end{subfigure}
    \begin{subfigure}[]{.445\textwidth}
        \small
        \resizebox{\linewidth}{!}{
        \begin{tabular}{@{\;}l@{\;}c@{\;\;}c@{\;}}
            \toprule
            & \# Original Answers & \# Entity Answers \\ \midrule
            VG & 50,130 & 3,460 \\
            OK-VQA & \pz4,214 & 1,600 \\
            Text-VQA & 19,500 & 3,562 \\
            VQA v2 & 26,748 & 4,337 \\
            Visual7W & \pz7,588 & 1,945 \\
            \bottomrule
        \end{tabular}
        }
    \end{subfigure}
    \caption{Number of unique entities on Entity split (left) and Query split (right). We compare it against the \# of entities before applying pre-processing. Note that VQA datasets contain massive non-entity answers, or collapsed answers, which leads to a large reduction in numbers after pre-processing.}
    \label{fig:unique_entity_distribution}
\end{figure*}

%% file: tables_and_figures/appendix/dataset_stats_appendix.tex
\tikzset{
  barlabels/.style={font=\footnotesize\sffamily},
  declare function={
    barheight=5pt;
  }
}

\begin{figure*}[htp]
    \centering

\begin{tikzpicture}[
  y=0.34cm,
  x=0.5cm,
]

\foreach [count=\i from 0] \p/\t in{
    19.4/animal,
    16.0/plant, 
    13.8/building,
    5.8/location,
    4.1/food,
    4.0/person,
    3.8/organization,
    1.9/vehicle,
    1.9/material,
    1.5/facility,
    1.0/sports,
    0.8/equipment,
    0.5/activity,
    25.5/others
}
  {
   \node [anchor=base east,
          barlabels,
          name=i-\i] at (0,-\i) {\t};
   \fill [blue!40] (i-\i.base east) rectangle ++(\p,barheight)  ++(0,-barheight)
          node[barlabels, 
               black,
               anchor=base west] {\p};
  }

\end{tikzpicture}
    \caption{Distribution of the entities in our datasets (Grouped by their super category).}
    \label{fig:dataset_stats_full}
\end{figure*}
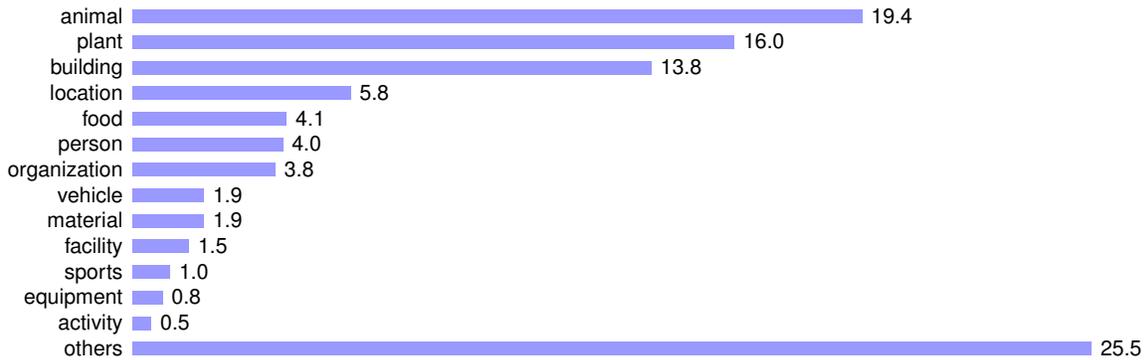

%% file: tables_and_figures/appendix/mixmore_annotation_interface.tex
\begin{figure*}[ht]
    \centering
    \includegraphics[trim={0 2cm 0 0},clip, width=\textwidth]{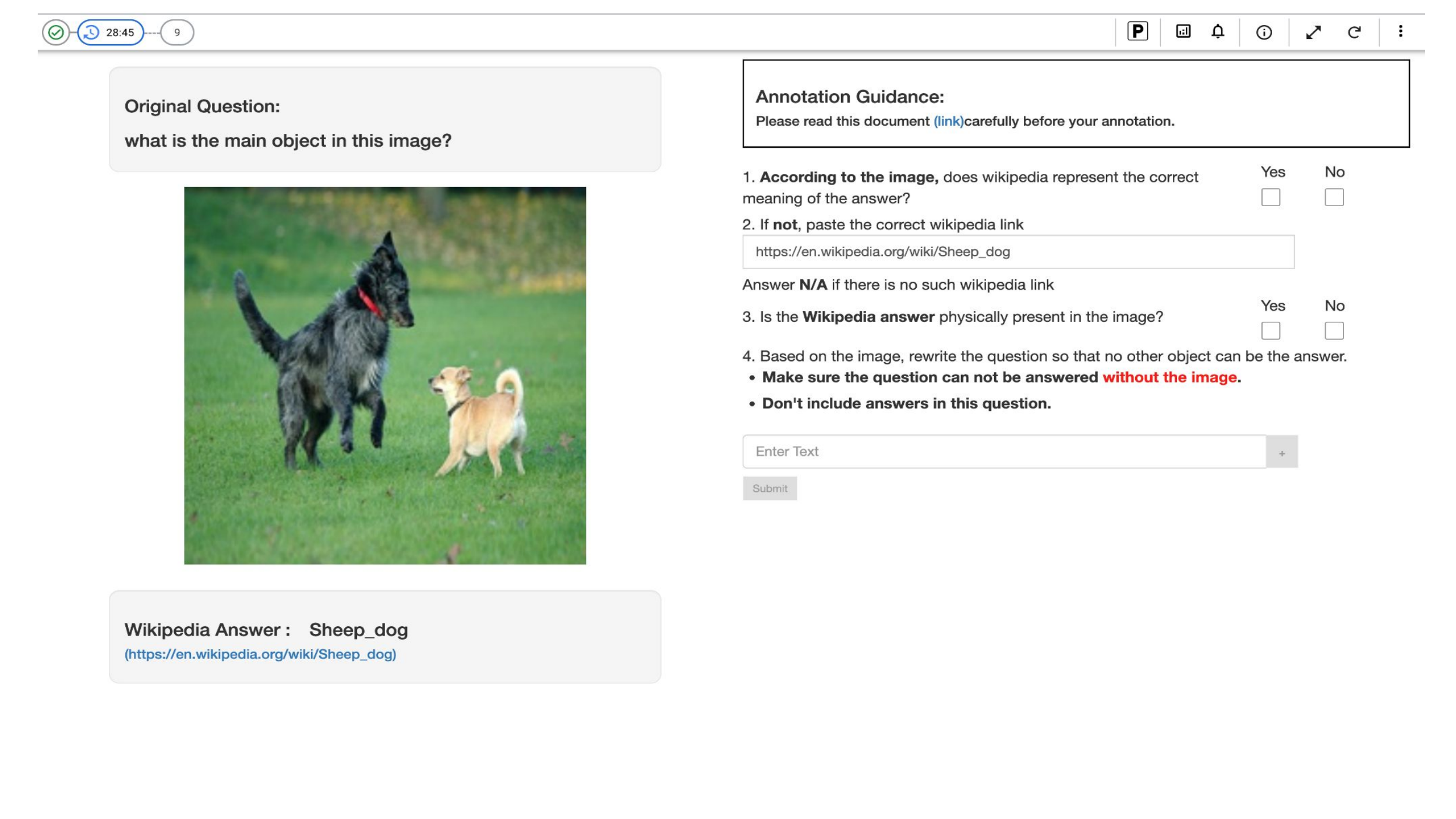}
    \caption{Annotation inferface}    
    \label{fig:annotation_interface}
\end{figure*}